%% file: main.tex
\documentclass[10pt,twocolumn,letterpaper]{article}

\usepackage[pagenumbers]{iccv} % To force page numbers, e.g. for an arXiv version


\definecolor{iccvblue}{rgb}{0.21,0.49,0.74}
\usepackage[pagebackref,breaklinks,colorlinks,allcolors=iccvblue]{hyperref}
\usepackage{multirow}
\usepackage{pifont}
\usepackage{makecell}
\usepackage{xcolor}

\definecolor{darkblue}{rgb}{0.0, 0.0, 0.5}

\def\paperID{13239} % *** Enter the Paper ID here
\def\confName{ICCV}
\def\confYear{2025}

\title{Memory-Efficient Optical Flow via Radius-Distribution Orthogonal Cost Volume}

\author{Gangwei Xu \quad Shujun Chen \quad Hao Jia \quad Miaojie Feng \quad Xin Yang\footnotemark[2]\\
[3mm]
Huazhong University of Science and Technology}

\begin{document}
\maketitle
\input{sec/0_abstract}

\renewcommand{\thefootnote}{\fnsymbol{footnote}}
\footnotetext[2]{Corresponding author.}

\input{sec/1_intro}

\input{sec/2_related}
\input{sec/3_method}
\input{sec/4_experiment}

\input{sec/5_conclusion}
{
    \small
    \bibliographystyle{ieeenat_fullname}
    \bibliography{main}
}

\end{document}

%% file: sec/0_abstract.tex
\begin{abstract}
Current optical flow methods based on dense all-pairs correlations or global matching achieve impressive performance. However, their memory consumption increases quadratically with input resolution, rendering them impractical for high-resolution images or hardware-constrained applications. In this paper,
we present MeFlow, a novel memory-efficient method for high-resolution optical flow estimation. The key of MeFlow is a recurrent local orthogonal cost volume representation, which decomposes the 2D search space dynamically into two 1D orthogonal spaces, enabling our method to scale effectively to very high-resolution inputs. To preserve essential information in the orthogonal space, we utilize self attention to propagate feature information from the 2D space to the orthogonal space. We further propose a radius-distribution multi-scale lookup strategy to model the correspondences of large displacements at a negligible cost. We verify the efficiency and effectiveness of our method on the challenging Sintel and KITTI benchmarks, and real-world 4K ($2160\!\times\!3840$) images. Our method achieves competitive performance on both Sintel and KITTI benchmarks, while maintaining the highest memory efficiency on high-resolution inputs. Code: \textcolor{magenta}{https://github.com/gangweiX/MeFlow}.
\end{abstract}

%% file: sec/1_intro.tex
\section{Introduction}
\label{sec:intro}

Optical flow, which estimates per-pixel 2D motion between successive two images, is a fundamental task for many applications such as action recognition~\cite{sevilla2019integration, sun2018actor}, video processing~\cite{yang2021self, yang2021learning, hdrflow} and autonomous driving~\cite{kitti15, promptingdepth,zhang2024leveraging}. Despite decades of research, accurately estimating optical flow with high memory and computational efficiency, in particular for high-resolution images (more than 4K resolution), remains very challenging.

State-of-the-art optical flow estimate methods~\cite{flownet, pwc-net,raft} mainly leverage deep learning frameworks with an essential component called cost volume. The pioneering deep learning-based optical flow method FlowNetC~\cite{flownet}, employs convolutional neural networks (CNNs) to extract feature maps. It constructs a cost volume by computing correlations between each pixel in the source image and its potential correspondences within a 2D search window of the target image. After that, relative pixel displacements are inferred and regressed from the cost volume via CNNs. Motivated by the success of FlowNetC, a serial of methods~\cite{flownet2,pwc-net,liteflownet,raft} have been proposed and achieved the state-of-the-art performance.

\begin{figure}
	\centering
	\includegraphics[width=1.0\linewidth]{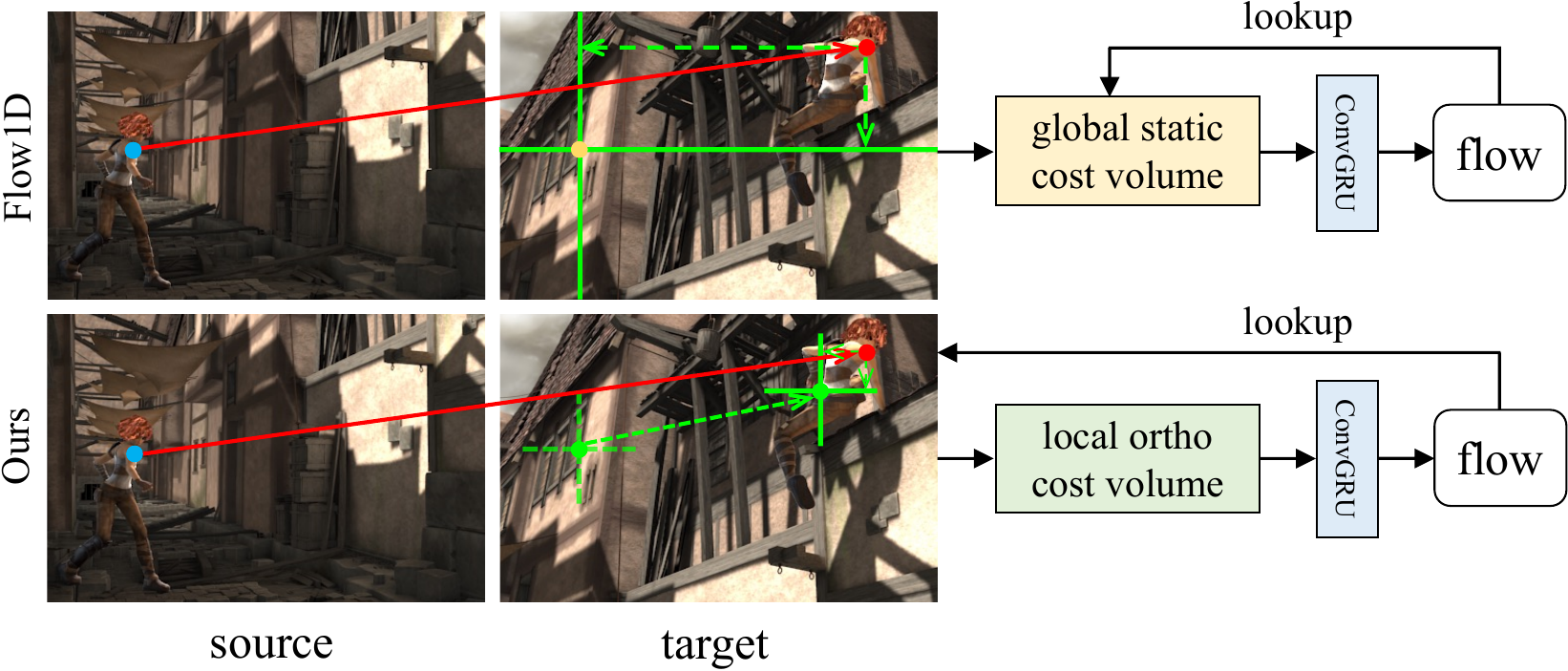}
	\caption{Comparison of our local orthogonal cost volume and the global static cost volume in Flow1D~\cite{flow1d}. Flow1D only searches two static orthogonal lines ({\color{green}green lines}) for every pixel in source image. In contrast, our method can dynamically search the entire 2D space based on the current updated flow ({\color{green}green point}).}
	\label{fig:lov}
 \vspace{-10pt}
\end{figure}

% \begin{figure}
% 	\centering	\includegraphics[width=1.0\linewidth]{figures/rank_new_crop.pdf}
%     \vspace{-18pt}
% 	\caption{Comparison with state-of-the-art optical flow methods. \textbf{Left:} GPU memory consumption on 1080p ($1080\!\times\!1920$) resolution inputs. \textbf{Right:} Results on Sintel and KITTI benchmarks. Our MeFlow achieves the highest memory efficiency and competitive accuracy. The red cross {\color{red}$\times$} denotes that FlowFormer~\cite{flowformer} causes an out-of-memory error on our 24G GPU.}
% 	\label{fig:ranking}
%  \vspace{-10pt}
% \end{figure}

\begin{figure}
	\centering
	\includegraphics[width=1.0\linewidth]{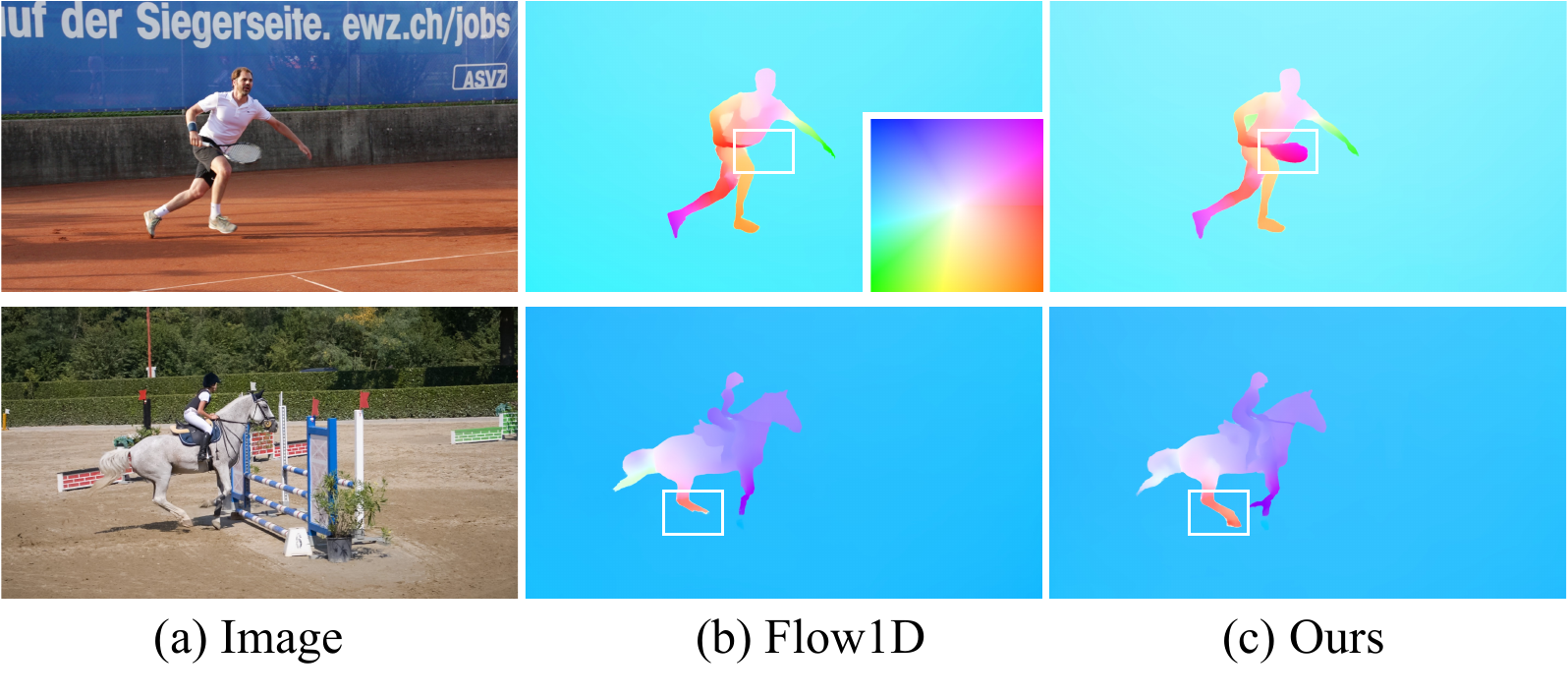}
	\caption{Comparisons with Flow1D~\cite{flow1d} on DAVIS dataset. Flow1D struggles to handle large motions. In contrast, our method performs well in these regions. Flow visualization is based on the color wheel shown on the corner of the first flow map.}
	\label{fig:large_motion}
 \vspace{-10pt}
\end{figure}

A recent and notable work is RAFT~\cite{raft} which computes all-pairs correlations to form a full 4D cost volume. 
% Local cost values are indexed iteratively from the full cost volume and used to update flow estimate results via Convolutional Gated Recurrent Unit (ConvGRU).
As the full 4D cost volume can encode complete pairwise matching information, RAFT achieves impressive performance on several established benchmarks. However, the complexity of the 4D cost volume, i.e., $\mathcal{O} (H\!\times \! W\!\times \!H\!\times \!W)$, grows quadratically with input image resolution, preventing its application to very high-resolution images.
To reduce the memory cost, SCV~\cite{scv} proposes a sparse cost volume that only stores the top-k correlations for each pixel. However, this approach may miss true matches for textureless and ambiguous regions, and in turn yield a 21.0\% performance degradation on KITTI compared to RAFT. Subsequently, Flow1D~\cite{flow1d} constructs two 3D cost volumes to replace RAFT’s 4D cost volume. To construct two 3D cost volumes, Flow1D decomposes a 2D search space into two orthogonal 1D search spaces by attention, as shown in \cref{fig:lov}. However, global attention in Flow1D is position-sensitive, yielding miss detection of true matches with large motions. As a result, Flow1D struggles to handle large motions, shown in \cref{fig:large_motion} and \cref{tab:large_motion}.
% The work most similar to ours is Flow1D~\cite{flow1d} which decomposes a 2D search space into two orthogonal 1D search spaces via 1D attention and 1D correlation. Specifically, Flow1D leverages 1D horizontal (vertical) self attention to make every pixel of the source image encode the information of its entire row (column). Next, 1D vertical and horizontal cross attention are performed between the source and target images respectively for global feature propagation. Finally, two 3D cost volumes are constructed by 1D correlation. However, it is challenging to ensure that every point is aware of all other pixels within the same row and column, particularly for dissimilar pixels. As a result, the global propagation in Flow1D is position-sensitive, yielding miss detection of true matches with large displacements.

In this paper, we introduce a novel memory-efficient network for high-resolution optical flow estimation, named MeFlow. The core of MeFlow is a novel Local Orthogonal Cost Volume (LOV) that decomposes the 2D search space dynamically into two 1D orthogonal spaces. Unlike Flow1D which constructs global static cost volume, our LOV is a local dynamic cost volume that is built recurrently based on the current updated flow (shown in \cref{fig:lov}).

Our LOV is constructed via the following three steps (\cref{fig:network}). First, we apply 1D local attention to the target feature to generate the vertically attended feature and horizontally attended feature. Since a pixel is more likely to belong to the same category as its neighboring pixels than those far away, this local attention can propagate reliable and effective information to the surroundings. Second, in each iteration, we index vertically attended feature and horizontally attended feature along the orthogonal direction based on the current flow estimation. Finally, we perform 1D correlation between these flow-indexed attended target feature and the source image feature to construct LOV, which can model 2D space correspondences.

% with search radius based on the current flow estimation along the horizontal direction, and horizontally attended target features along the vertical direction. With these flow-indexed attended features, we can construct a dynamic orthogonal cost volume via 1D correlation, which can model 2D space correspondences. 

Different from RAFT's multi-scale 4D cost volume and Flow1D's two 3D cost volumes that only need to be constructed once before GRU iteration, our local orthogonal cost volume needs to be built for each GRU iteration. Therefore, properly setting the value of search radius $R$ is crucial for achieving both accuracy and efficiency. Increasing $R$ can model large motions, but it also increases the computational cost. To address this issue, we propose a novel radius-distribution multi-scale (RDMS) orthogonal lookup strategy (\cref{fig:rdms}). We index finer-resolution (e.g., 1/8) attended target features within a smaller search range and coarser-resolution (e.g., 1/16 and 1/32) features in the outer search region. The proposed RDMS strategy can handle large motions with less computational cost compared to RAFT's multi-scale 4D cost volume (\cref{fig:rdms}).

The complexity of our LOV is $\mathcal{O} (H\!\times\!W\!\times\!(4R\!+\!2))$, where $R$ (default value is 8) denotes radius-distribution lookup radius. In comparison, the complexity of the cost volume in RAFT and Flow1D is $\mathcal{O} (H\!\times \! W\!\times \!H\!\times \!W)$ and $\mathcal{O} (H\!\times \! W\!\times \!(H\!+\!W))$ respectively. For high-resolution inputs, $4R\!+\!2 \ll H\!+\!W \ll H\!\times\!W$ indicates that our LOV has a smaller peak memory usage in inference compared to Flow1D and RAFT.

% The complexity of a 4D cost volume in RAFT\cite{raft} is $\mathcal{O} (H\!\times \! W\!\times \!H\!\times \!W)$, and the complexity of two 3D cost volumes in Flow1D\cite{flow1d} is $\mathcal{O} (H\!\times \! W\!\times \!(H\!+\!W))$. In contrast, the complexity of our dynamic orthogonal cost volume is $\mathcal{O} (H\!\times\!W\!\times\!(4R\!+\!1))$, where $R$ (default 8) denotes multi-scale lookup radius, and $(4R\!+\!1) \ll H\!+\!W \ll H\!\times\!W$ for high-resolution inputs.

We demonstrate the efficiency and effectiveness of our approach on the challenging datasets including Sintel~\cite{sintel}, KITTI~\cite{kitti15}, as well as the high-resolution DAVIS datasets~\cite{davis17,davis19}. Our approach consumes $6 \times$ less memory than RAFT on 1080p resolution images and can handle real-world 4K images with only 5.4G memory usage. RAFT can not handle 4K images due to causing an out-of-memory error on our 24G GPU. Meanwhile, we achieved better results than RAFT on the KITTI benchmark. Our method also outperforms almost all memory-efficient methods, e.g., exceeding Flow1D by 21.1\% and SCV by 19.8\% on the KITTI benchmark.

Our major contributions can be summarized as follows:
\begin{itemize}
	\item We propose a novel local orthogonal cost volume, which dynamically encodes the correspondences in 2D space into two 1D orthogonal spaces, enabling our method to scale well to very high-resolution inputs.
        \item We propose a new radius-distribution multi-scale lookup strategy to efficiently model the correspondences of large displacements.
        % , which is different from RAFT's multi-scale cost volumes.
	\item We design a memory-efficient flow network, named MeFlow, which achieves the highest memory efficiency among all published methods and outperforms almost all memory-efficient methods on the challenging Sintel and KITTI benchmarks.
\end{itemize}

%% file: sec/2_related.tex
\section{Related Work}
\label{sec:related}

\paragraph{Deep Flow Methods} Deep convolutional neural networks have shown great success in optical flow estimate tasks. FlowNet~\cite{flownet} is an pioneer end-to-end CNN-based network for optical flow, which inspires a serial of subsequent works~\cite{flownet2,pwc-net,liteflownet}. FlowNet2~\cite{flownet2} introduces a warping operation and stacks multiple basic models to improve model capacity and performance. To further reduce model size and inference time, several methods~\cite{pwc-net+,liteflownet2,irr-pwc} propose the coarse-to-fine cost volume pyramid. However, these methods tend to miss fast-moving small objects in the coarse stage. To address this issue, RAFT~\cite{raft} proposes to construct an all-pairs 4D cost volume, recurrently indexing the cost values and optimizing them in the convolution GRU block to update optical flow. RAFT achieves the state-of-the-art performance on optical flow benchmarks. Based on the RAFT architecture, some recurrent methods~\cite{gma, kpaflow, gmflownet, craft, agflow, igev, igev++, selectivestereo, monster, gaflow,flowformer++,videoflow} have been proposed to further improve the accuracy. For example, GMA~\cite{gma} and KPAFlow~\cite{kpaflow} aggregate global or local motion features via attention mechanism to resolve ambiguities caused by occlusions, yielding the state-of-the-art performance. However, due to the use of the 4D cost volume ($H \!\times \!W \! \times\!H \!\times \!W$), the memory consumption of these two methods grows quadratically with the input resolution, which makes them very difficult to scale to high-resolution images. 
\vspace{-10pt}

% In contrast, we present a memory-efficient orthogonal cost volume, enabling a good scalability of our method to high-resolution images.

\paragraph{Memory-Efficient Flow Methods} Several recent works have designed efficient cost volume representations~\cite{flow1d, scv, hcvflow} to replace the original 4D cost volume in RAFT~\cite{raft} for processing high-resolution images. For example, SCV~\cite{scv} constructs a sparse cost volume by computing the top-k correlations for each pixel. Flow1D~\cite{flow1d} constructs two 3D cost volumes in the vertical and horizontal directions respectively to represent 4D spatial correlation. Even though they improve memory efficiency, they suffer from nontrivial loss of accuracy compared to the original 4D cost volume in RAFT. DIP~\cite{dip} introduces the idea of the conventional method Patchmatch into the end-to-end optical flow network. Benefiting from propagation and local search of Patchmatch which can effectively avoid the 4D cost volume construction, DIP can achieve high-precision results with a low memory cost. However, sequential ConvGRU modules (i.e., alternating the propagation block and the local search block) in DIP are time-consuming, yielding a $2.4 \times$ slower speed than our method for 1080p ($1080\times1920$) resolution images. Among these memory-efficient methods~\cite{scv,flow1d,dip}, our method achieves the highest efficiency in memory while achieving the state-of-the-art performance on the KITTI benchmark.

\paragraph{Attention mechanism} Attention has achieved great success in many vision tasks~\cite{swintrans, dualatt,ccnet,axial,detectiontrans,loftr,gma,acvnet,gmflow,xu2022accurate} due to its ability to model long-range dependencies. For example, CCNet~\cite{ccnet} proposes efficient criss-cross attention to aggregate contextual information. GMA~\cite{gma} uses attention to aggregate global motion features, and GMFlow~\cite{gmflow} exploits the cross attention to obtain discriminative features for matching. Flow1D~\cite{flow1d} uses 1D cross attention for global feature propagation and then computes 1D correlation. However, such global propagation in Flow1D is position sensitive, yielding miss detection of true matches with large displacements. Instead, the local attention in our MeFlow can propagate more reliable and effective information.

%% file: sec/3_method.tex
\section{MeFlow}
\label{sec:method}

\begin{figure*}[t]
    \centering
    \includegraphics[width=1.0\linewidth]{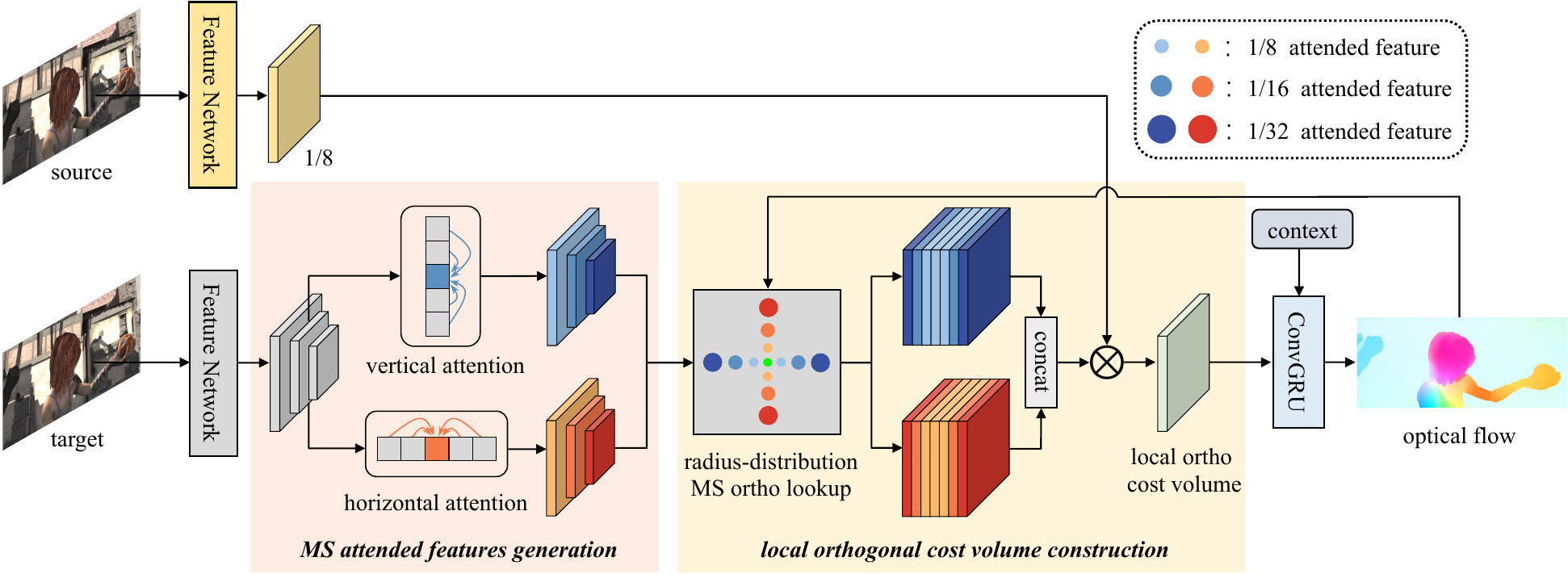}
    \caption{Overview of the proposed MeFlow. We apply vertical and horizontal attention to the multi-scale target features to generate multi-scale attended features. Then, in each iteration, we index the multi-scale vertically attended features and horizontally attended features along the horizontal and vertical direction respectively based on the current updated flow ({\color{green}green point}). Specially, the proposed radius-distribution multi-scale (MS) orthogonal lookup can index finer-resolution features at small radius and coarser-resolution at large radius. Finally, we construct a dynamic orthogonal cost volume by performing 1D correlation between the source image feature and the dynamically indexed attended features.}
    
    % which We apply vertical and horizontal attention to the multi-scale target features to generate multi-scale attended features. Then, in each iteration, we index the multi-scale vertically attended features and horizontally attended features along the horizontal and vertical direction respectively based on the current updated flow ({\color{green}green point}). Finally, we construct a local orthogonal cost volume by performing 1D correlation between the source image feature and the dynamically indexed attended features.}
    \label{fig:network}
    % \vspace{-10pt}
\end{figure*}

\begin{figure}
\centering
\includegraphics[width=1.0\linewidth]{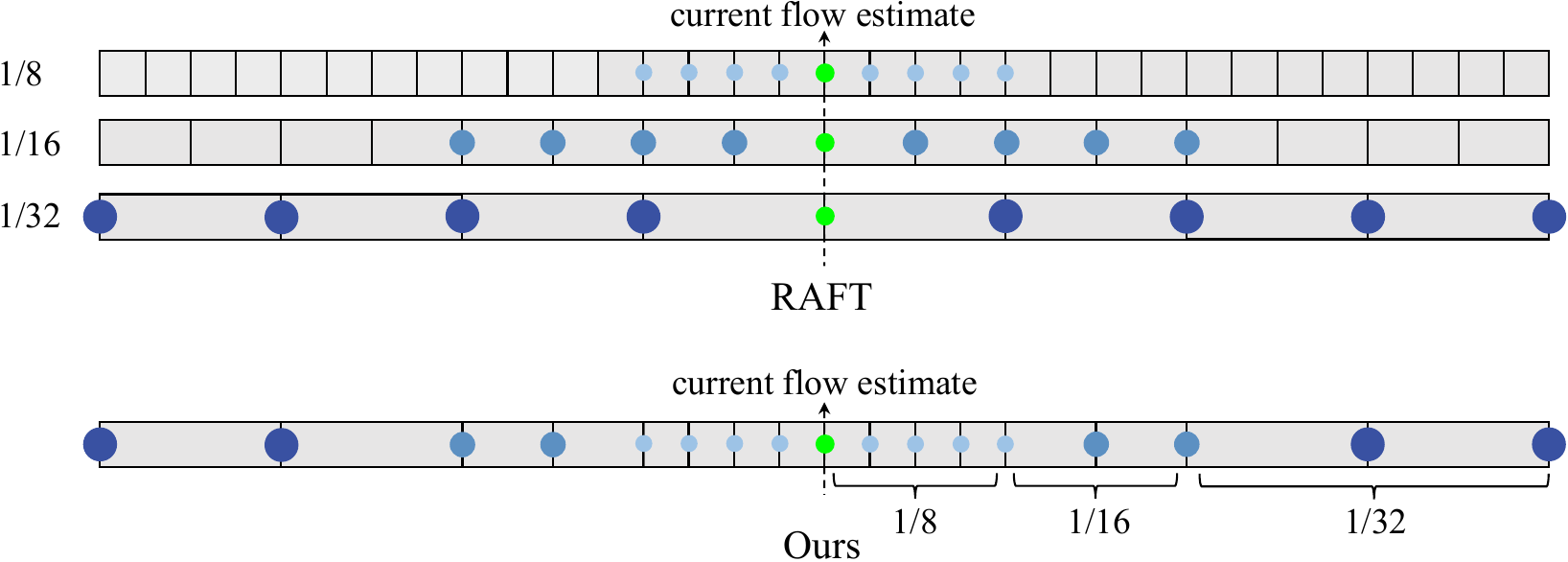} % Reduce the figure size so that it is slightly narrower than the column.
\caption{RAFT's multi-scale strategy v.s. our radius-distribution multi-scale strategy.}
\label{fig:rdms}
\end{figure}
Optical flow is essentially a 2D search problem, which has the quadratic
complexity with respect to the 2D search window. Thus, directly searching in the 2D space could become computationally intractable for large displacements, limiting its application to high-resolution images which are ubiquitous in practical scenarios due to the popularity of high-definition cameras.
 
This section presents MeFlow, which can handle high-resolution images with a low memory cost by decomposing the 2D search space dynamically into two 1D orthogonal spaces. Our MeFlow mainly consists of two components: 1) Multi-scale Attended Features Generation and 2) Local Orthogonal Cost Volume Construction. Next, we will introduce the details of each component.

\subsection{Multi-Scale Attended Features Generation}
Given source and target images ${\textbf I}_s$ and ${\textbf I}_t$, we first extract $8 \times$ downsampled features ${\textbf F}_s, {\textbf F}_t \in \mathbb{R}^{H \times W \times D}$ with a weight-sharing feature network, where $H, W$ and $D$ denote height, width and feature dimension, respectively. Following RAFT~\cite{raft}, the feature network consists of 6 residual blocks, 2 at 1/2 resolution, 2 at 1/4 resolution, and 2 at 1/8 resolution. 

To produce multi-scale attended target features, we apply one and two average pooling operations with a $2 \times 2$ kernel size to the target feature ${\textbf F}_{t}$ to obtain 1/16 resolution target feature ${\textbf F}_{t}^{d1}$ and 1/32 resolution target feature ${\textbf F}_{t}^{d2}$ respectively. Then, we use vertical and horizontal attention (\cref{fig:network}) on the these generated target features (i.e. ${\textbf F}_{t}$, ${\textbf F}_{t}^{d1}$, ${\textbf F}_{t}^{d2}$) to obtain vertically attended target features $\hat{\textbf F}_{t,v}$, $\hat{\textbf F}_{t,v}^{d1}$, $\hat{\textbf F}_{t,v}^{d2}$ and horizontally attended target features $\hat{\textbf F}_{t,h}$
$\hat{\textbf F}_{t,h}^{d1}$, $\hat{\textbf F}_{t,h}^{d2}$.
% Then, we perform vertical and horizontal attention respectively on the target feature ${\textbf F}_t$ to generate two new features, $\hat{\textbf F}_{t, v}$ and $\hat{\textbf F}_{t, h}$ (\cref{fig:network}). 
In vertically attended target features, every feature at a point ${\textbf P}$ contains feature information of a set of points which lie on the same column of ${\textbf P}$. Similarly, in horizontally attended target features, every feature at a point ${\textbf P}$ contains feature information of a set of points which lie on the same row of ${\textbf P}$. Here, we detail the generation process of $\hat{\textbf F}_{t, v}$, and the generation process of other attended target features is similar. 

Given the target feature ${\textbf F}_t$, we first apply two convolutional layers with $1\!\times\!1$ filters to ${\textbf F}_t$ to generate two feature maps ${\textbf Q}$ and ${\textbf K}$, respectively, where ${\textbf Q}$, ${\textbf K} \in \mathbb{R}^{H \times W \times D}$. 
% Then we generate attention maps ${\textbf A} \in \mathbb{R}^{(2R+1) \times H \times W}$ by affinity operation. 
Then we unfold ${\textbf K}$ with radius $R_0$ along the vertical direction, and the unfolded ${\textbf K}$ is denoted as ${\textbf K}_{u} \in \mathbb{R}^{(2R_0+1) \times H \times W \times D}$. We generate correlation matrix ${\textbf M} \in \mathbb{R}^{(2R_0+1) \times H \times W}$ by affinity operation,
\begin{equation}
\begin{split}
\label{eq:linear}
{\textbf M} (h, w) & = \frac{{\textbf Q} (h,w) {{\textbf K}_u^T} (h,w)}{\sqrt{D}}  \in \mathbb{R}^{2R_0+1}, \\
\end{split}
\end{equation}
where $\frac{1}{\sqrt{D}}$ is a normalization factor to avoid large values. Then, we normalize the first dimension of ${\textbf M}$ with the softmax to obtain attention maps ${\textbf A} \in \mathbb{R}^{(2R_0+1) \times H \times W}$,
\begin{equation}
\begin{split}
\label{eq:softmax}
{\textbf A} & = softmax ({\textbf M}).
\end{split}
\end{equation}

Finally, the attended target feature $\hat{\textbf F}_{t, v}$ is computed by vertically aggregating ${\textbf F}_t$,
\begin{equation}
\begin{split}
\label{eq:softmax}
\hat{\textbf F}_{t, v} (h, w) & = \sum_{r = -R_0}^{R_0}{\textbf A}(r, h, w) {\textbf F}_t (h + r, w).
\end{split}
\end{equation}

% To produce multi-scale attended target features, we apply one and two average pooling operations with a $2 \times 2$ kernel size to the target feature ${\textbf F}_{t}$ to obtain 1/16 resolution target feature ${\textbf F}_{t}^{1}$ and 1/32 resolution target feature ${\textbf F}_{t}^{2}$ respectively. Then, we use vertical and horizontal attention again on the these pooled target features to obtain attended features $\hat{\textbf F}_{t,v}^{1}$, $\hat{\textbf F}_{t,v}^{2}$, $\hat{\textbf F}_{t,h}^{1}$ and $\hat{\textbf F}_{t,h}^{2}$.
\begin{figure*}
\centering
\includegraphics[width=0.95\textwidth]{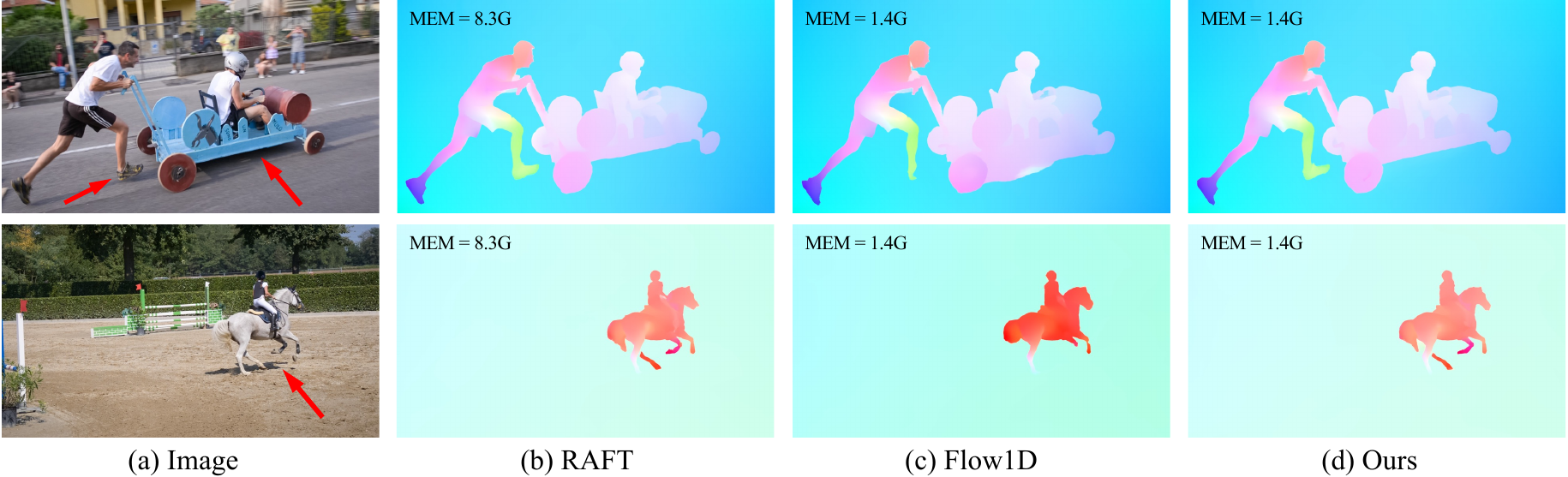} % Reduce the figure size so that it is slightly narrower than the column.
\caption{Comparisons on high-resolution ($1080 \times 1920$) images from DAVIS. We achieve comparable results with RAFT~\cite{raft} while consuming $6 \times$ less memory. We achieve more accurate results than the SOTA memory-efficient Flow1D~\cite{flow1d} (pointed by red arrows).}
\label{fig:comp_2k}
\vspace{-10pt}
\end{figure*}

\subsection{Local Orthogonal Cost Volume Construction}
For each GRU iteration, we use the current optical flow to index a subset of multi-scale attended target features. Subsequently, we perform 1D correlation between these indexed target features and the source feature to construct a local orthogonal cost volume. Different from RAFT's multi-scale 4D cost volume and Flow1D's two 3D cost volumes that only need to be constructed once, our local orthogonal cost volume needs to be built for each GRU iteration. Thus, we propose a novel radius-distribution multi-scale strategy, which is more efficient than RAFT's multi-scale strategy, shown in \cref{fig:rdms}. 
Next, we will present the details of radius-distribution multi-scale strategy.

For a pixel at the position $(h, w)$ on the source image feature ${\textbf F}_s$, we find its corresponding position $(h^{\prime}, w^{\prime})$ on the attended target image feature based on the current optical flow estimate ${\textbf f} = (f_x, f_y)$, i.e. $(h^{\prime}, w^{\prime}) = (h+f_y, w+f_x)$. Then, centered on the pixel position $(h^{\prime}, w^{\prime})$, we index the multi-scale attended target features along the horizontal and vertical directions respectively. Specially, for the attended target features at 1/8, 1/16 and 1/32 resolution, the lookup radius is $R_0=4$, $R_1=2$ and $R_3=2$ respectively, and the total lookup radius is $R = R_0+R_1+R_2$ (\cref{fig:rdms}). For example, with vertically attended target features $\hat{\textbf F}_{t,v}$, $\hat{\textbf F}_{t,v}^{d1}$ and $\hat{\textbf F}_{t,v}^{d2}$, we can calculate the horizontal cost volume ${\textbf C}_h$ along the horizontal direction,
\begin{equation}
\begin{split}
\label{eq:linear}
{\textbf C}_h^{d0} (R\!+\!r_0, h, w) & \!= \!\frac{1}{\sqrt{D}} {\textbf F}_s(h, w) \boldsymbol{\cdot} \hat{\textbf F}_{t,v} (h^{\prime}, w^{\prime}\!+\!r_0), \\
{\textbf C}_h^{d1} (R\!+\!r_1\!/\!2, h, w) & \!= \!\frac{1}{\sqrt{D}} {\textbf F}_s(h, w) \boldsymbol{\cdot} \hat{\textbf F}_{t,v}^{d1} (h^{\prime}\!/\!2, w^{\prime}/2\!+\!r_1\!/\!2), \\
{\textbf C}_h^{d2} (R\!+\!r_2\!/\!4, h, w) & \!= \!\frac{1}{\sqrt{D}} {\textbf F}_s(h, w) \boldsymbol{\cdot} \hat{\textbf F}_{t,v}^{d2} (h^{\prime}\!/\!4, w^{\prime}/4\!+\!r_2\!/\!4), \\
{\textbf C}_h & = \text{Concat}\{{\textbf C}_h^{d0}, {\textbf C}_h^{d1}, {\textbf C}_h^{d2}\},
\end{split}
\end{equation}
where $r_0\!\in\!\{ -4, -3, -2, -1, 0, 1, 2, 3, 4\}$, $r_1\!\in\!\{ -8, -6, 6, 8\}$, $r_2\!\in\!\{ -16, -12, 12, 16\}$, $\boldsymbol{\cdot}$ denotes the dot product, ${\textbf C}_h \in \mathbb{R}^{(2R+1) \times H \times W}$.

\begin{figure}
	\centering
	\includegraphics[width=1.0\linewidth]{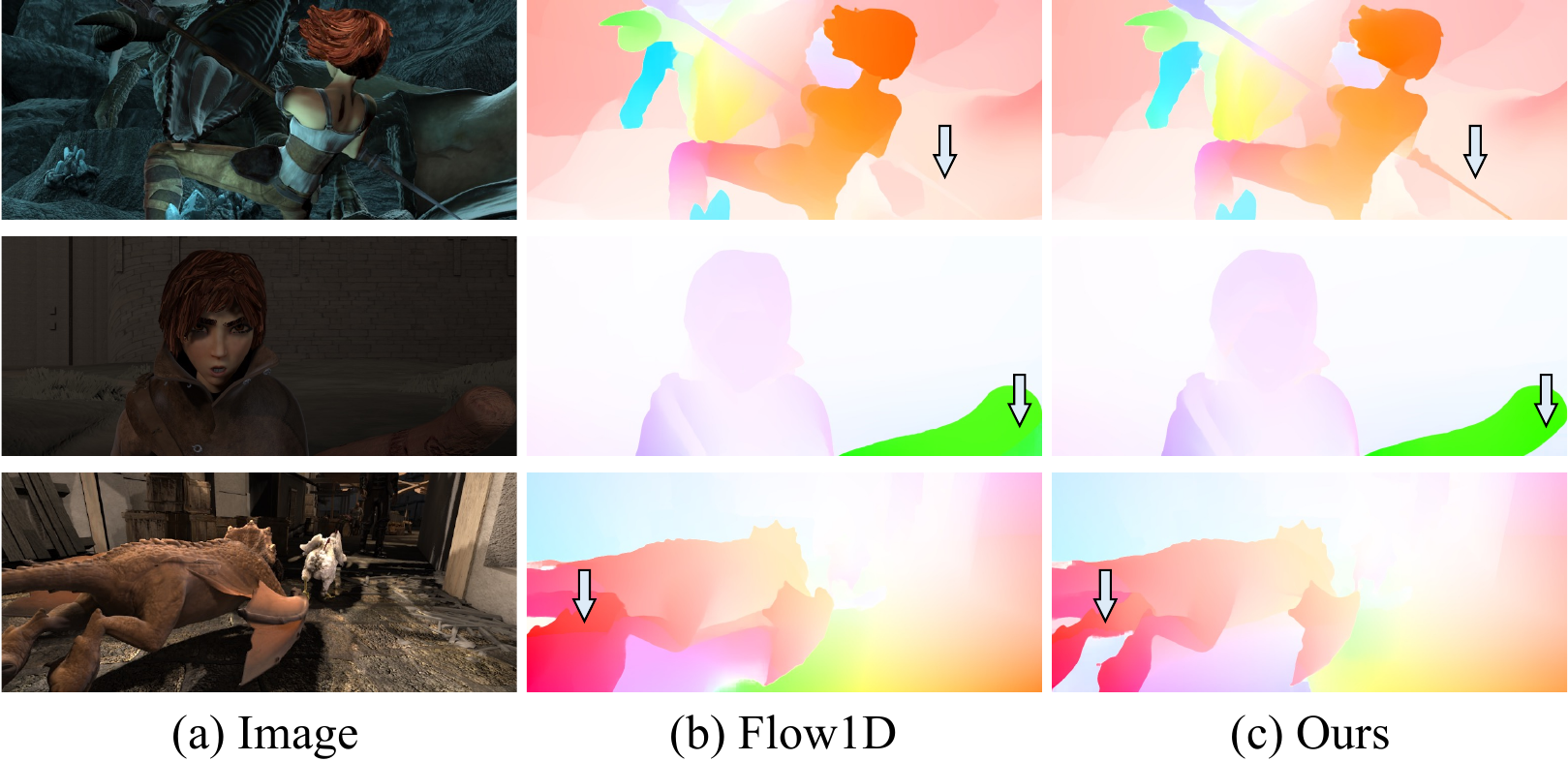}
	\caption{Qualitative comparisons on the Sintel test set. Our method can preserve more fine structure than Flow1D.}
 % Significant improvements are highlighted by arrows.}
	\label{fig:sintel}
 \vspace{-5pt}
\end{figure}

\begin{table*}
% \footnotesize
% \setlength{\tabcolsep}{2pt}
\caption{Ablation study. OC denotes orthogonal correlation, OA denotes orthogonal attention (i.e., vertical and horizontal attention), and RDMS denotes radius-distribution multi-scale lookup strategy. Base represents RAFT's alternate implementation (lower memory cost). All models are trained on FlyingChairs and FlyingThings3D. The final model, OC+OA+RDMS, is denoted as MeFlow.}
    \centering
    \begin{tabular}{lcccccccc}
    \toprule
    \multirow{2}{*}{Model} & \multicolumn{2}{c}{Sintel (train)} & \multicolumn{2}{c}{KITTI (train)}   & \multicolumn{2}{c}{$448 \times 1024$} & \multicolumn{2}{c}{$1080 \times 1920$}\\
    \cmidrule(lr){2-3} \cmidrule(lr){4-5} \cmidrule(lr){6-7} \cmidrule(lr){8-9}
    & Clean & Final & EPE  & F1-all  & \makecell{Memory \\(G)} & \makecell{Time \\(ms)} & \makecell{Memory \\(G)} & \makecell{Time \\(ms)} \\
    \midrule
    % Base (multi-scale, $(2r\!+\!1)^2\!\times\!3$) & { 1.45} & {2.74} & { 4.83} & { 16.94} & 5.22 & 0.66 & 108 & 2.90 & 545 \\
    Base ($(2r\!+\!1)^2$) & 1.55 & 2.88 & 5.63 & 18.26  & 0.65 & 68 & 2.81 & 350 \\
    % \midrule
    OC ($(2r\!+\!1)\!\times\!2$) & 1.68 & 2.92 & 6.08 & 18.94 & 0.33 & 52 & 1.39 & 220 \\
    OC \!+\! OA & 1.56 & 2.82 & 5.69 & 18.21 & 0.33 & 52 & 1.39 & 220 \\
    OC \!+ \!OA \!+\! RDMS & \textbf{1.49} & \textbf{2.75} & \textbf{5.31} & \textbf{16.65} & 0.33 & 65 & 1.39 & 260 \\
    \bottomrule
    \end{tabular}
    \label{tab:ablation}
\end{table*}

Similar to horizontal cost volume $\textbf {C}_h$, we perform vertical correlation on horizontally attended target features $\hat{\textbf F}_{t,h}$, $\hat{\textbf F}_{t,h}^{d1}$ and $\hat{\textbf F}_{t,h}^{d2}$ to obtain the vertical cost volume ${\textbf C}_v \in \mathbb{R}^{(2R+1) \times H \times W}$. Finally, we obtain the local orthogonal cost volume ${\textbf C}_o \in \mathbb{R}^{(4R+2) \times H \times W}$ by concatenating ${\textbf C}_h$ and ${\textbf C}_v$,
\begin{equation}
\begin{split}
\label{eq:linear}
{\textbf C}_o & = \text{Concat}\{{\textbf C}_h, {\textbf C}_v\}.
\end{split}
\end{equation}

The local orthogonal cost volume $\textbf {C}_o$ can model efficiently multi-scale 2D correspondences. Compared to RAFT's alternate implementation, which needs to search 243 (i.e. $(2\times4+1)^2 \times 3$) target feature vectors for each position on the source image, our method only needs to search 34 (i.e. $4 \times 8 + 2$) target feature vectors to model correspondences within the same scope, which is only 1/7 of RAFT (see \cref{fig:rdms}).

\subsection{Network Architecture}
\cref{fig:network} presents the architecture of our MeFlow. Following RAFT~\cite{raft}, we extract $8 \times$ downsampled features for source and target image by a feature network, and a context feature is also extracted from the source image with an additional context network. For brevity, we omit the context network. Next, we obtain target image features at three resolutions (i.e. 1/8, 1/16 and 1/32 resolution) through pooling operations, and perform vertical and horizontal attention on them respectively to generate multi-scale vertically attended features and horizontally attended features. In each iteration, we index multi-scale attended features along an orthogonal direction with the current flow estimation. Then we construct the local orthogonal cost volume by performing 1D correlation between the source image feature and the dynamically indexed attended target features. The local orthogonal cost volume, together with the current flow estimation and context feature, is then fed to a ConvGRU to update the current flow estimation. 

\begin{table*}
% \small
\caption{Comparisons with existing representative
methods. Memory and inference time are measured for $448\times 1024$ and $1080 \times 1920$ resolutions on our RTX 3090 GPU, and the GRU-based iteration numbers are 12 for RAFT, Flow1D and our MeFlow. {\bf Bold}: Best, \underbar{Underscore}: Second best.}
    \centering
    \begin{tabular}{lcccccccc}
    \toprule
    \multirow{2}{*}{Method} & \multicolumn{2}{c}{Sintel (train)} & \multicolumn{2}{c}{KITTI (train)} &\multirow{2}{*}{\begin{tabular}[x]{@{}c@{}}KITTI \\test \end{tabular}} &\multirow{2}{*}{\begin{tabular}[x]{@{}c@{}}Param\\(M) \end{tabular}}  & \multicolumn{2}{c}{Memory (G)} \\
    \cmidrule(lr){2-3} \cmidrule(lr){4-5} \cmidrule(lr){8-9}
    & Clean & Final & EPE  & F1-all & & & $448 \!\times \!1024$ & $1080\!\times \!1920$ \\
    \midrule
    RAFT~\cite{raft} & {\bf 1.43} & {\bf 2.71} & {\bf 5.04} & \underbar{17.40} & \underbar{5.10} & \underbar{5.26} & 0.48 &8.33 \\
    FlowNet2~\cite{flownet2} & 2.02 & 3.14 & 10.06 & 30.37 & 11.48 & 162.52 & 1.31 & 3.61 \\
    PWC-Net~\cite{pwc-net} & 2.55 & 3.93 & 10.35 & 33.67 & 9.80 & 9.37 & 0.86 & 1.57 \\
    Flow1D~\cite{flow1d} & 1.98 & 3.27 & 6.69 & 22.95 & 6.27 & 5.73 & \underbar{0.34} & \underbar{1.42} \\
    MeFlow (Ours) & \underbar{1.49} & \underbar{2.75} & \underbar{5.31} & {\bf 16.65} & {\bf 4.95} & {\bf 5.23} & {\bf 0.33} & {\bf 1.39} \\
    \bottomrule
    \end{tabular}
    \label{tab:compare_cost_volume}
    \vspace{-10pt}
\end{table*}

\subsection{Loss Function}
Following RAFT~\cite{raft}, we supervised our network using the $l_1$ loss between the predicted and ground truth flow over the full sequence of predictions, \{$\textbf{f}_1$,$\cdots$, $\mathbf{f}_N$\}, with exponentially increasing weights. Given ground truth flow $\mathbf{f}_{gt}$, the loss is defined as,
\begin{equation}
    \mathcal{L} = \sum_{i=1}^{N} \gamma^{N-i} ||\mathbf{f}_i-\mathbf{f}_{gt}||_1,
\end{equation}
where we set $\gamma=0.8$ in our experiments. 

\begin{figure}
	\centering
	\includegraphics[width=1.0\linewidth]{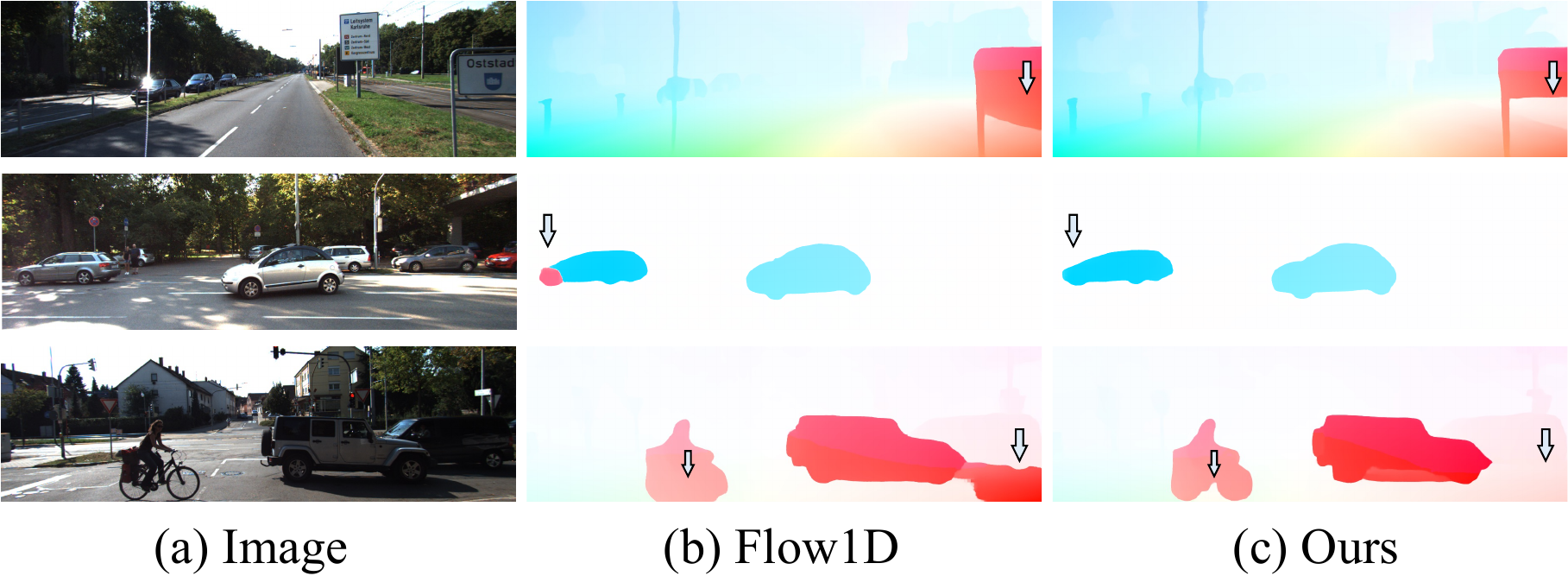}
	\caption{Qualitative comparisons on the KITTI test set. Our method performs well in large textureless regions.}
	\label{fig:kitti}
 \vspace{-5pt}
\end{figure}

%% file: sec/4_experiment.tex
\section{Experiments}
\label{sec:experiment}

\subsection{Experimental Setup}
\noindent\textbf{Datasets and evaluation setup.} We evaluate our MeFlow on the Sintel~\cite{sintel}, KITTI~\cite{kitti15} and high-resolution DAVIS~\cite{davis17,davis19} datasets. First, we pre-train our model on the FlyingChairs~\cite{flownet} and FlyingThings3D~\cite{flyingthings3d} datasets, and then evaluate the cross-dataset generalization ability on the Sintel and KITTI training sets. Second, we fine-tune our model on FlyingThings3D, Sintel, KITTI and HD1K, and then evaluate on the Sintel and KITTI test sets. For Sintel, we report the end-point-error (EPE) for evaluation. For KITTI, we report EPE and F1-all metrics as most comparison methods. F1-all represents the percentage of the outliers among all pixels. Finally, we evaluate the performance of our MeFlow on high-resolution (1080p and 4K) DAVIS images using the pre-trained model from Sintel to demonstrate that our method can scale up to very high-resolution images.

\noindent\textbf{Implementation details.} We implement our MeFlow in PyTorch and use AdamW~\cite{adamw} as the optimizer. Our convolutional backbone network is identical to RAFT’s model, except that our final feature dimension is 128, while RAFT’s is 256. Following the standard optical flow training procedure, we first train our model on FlyingChairs for 100K iterations with a batch size of 12 and then on FlyingThings3D for another 100K iterations with a batch size of 6. We then fine-tune our model on a combination of FlyingThings3D, Sintel, KITTI and HD1K for 100K iterations for Sintel evaluation and 50K iterations on KITTI for KITTI evaluation. We set the batch size to 6 for fine-tuning. We use 12 GRU-based iterations for training. For evaluation, we use 32 GRU-based iterations for Sintel and 24 GRU-based iterations for KITTI, respectively. For each iteration, the lookup radius $R$ on the multi-scale attended features is set to 8, which corresponds to 128 pixels in the original image resolution.

% \begin{table}[t]
% \small
%     \centering 
%     \begin{tabular}{lcccc}
%     \toprule
%     \multirow{2}{*}{Method }
%     &\multicolumn{2}{c}{Sintel (train)} & \multicolumn{2}{c}{KITTI (train)} \\
%     \cmidrule(lr){2-3} \cmidrule(lr){4-5}
%     & Clean & Final & EPE & F1-all \\
%     \midrule
%     $r$=2 & 1.64 & 2.85 & 5.73 & 18.69 \\ 
%     $r$=4 & 1.56 & 2.82 & 5.69 & 18.21 \\
%     % R=6 & 1.76 & 2.82 & 5.67 & 18.30 \\
%     $r$=8 & 1.55 & 2.81 & 5.45 & 17.92 \\
%     $R$=8 (MS) & {\bf 1.49} & {\bf 2.75} & {\bf 5.31} & {\bf 16.65} \\    
%     \bottomrule
%     \end{tabular}
%     \caption{Ablation study for orthogonal lookup radius.}
%     \label{tab:radius_r}
%     \vspace{-10pt}
% \end{table}

\begin{table}
\small
\caption{Comparisons with memory-efficient methods. Our method achieves the highest efficiency in memory.}
    \centering 
    \setlength{\tabcolsep}{2.pt} %
    \begin{tabular}{lccccc}
    \toprule
    \multirow{2}{*}{Method } &\multirow{2}{*}{\begin{tabular}[x]{@{}c@{}}KITTI\\test \end{tabular}}
    &\multicolumn{2}{c}{$448 \times 1024$} & \multicolumn{2}{c}{$1080 \times 1920$} \\
    \cmidrule(lr){3-4} \cmidrule(lr){5-6}
    & & Mem.(G) & Time(ms) & Mem.(G) & Time(ms) \\
    \midrule
    % RAFT & 5.10 & 0.48 & 64 & 8.33 & 300 \\ 
    SCV~\cite{scv} & 6.17 & 0.59 & 280 & 2.66 & 900 \\
    DIP~\cite{dip} & \textbf{4.21} & 0.67 & 180 & 2.90 & 620 \\
    Flow1D~\cite{flow1d} & 6.27 & \underline{0.34} & 52 & \underline{1.42} & 200 \\
    MeFlow (Ours) & \underline{4.95} & \textbf{0.33} & 65 & \textbf{1.39} & 260 \\    
    \bottomrule
    \end{tabular}
    \label{tab:comp_mem_effi}
    \vspace{-10pt}
\end{table}

\subsection{Ablation Study}
We perform ablation studies to validate the effectiveness and efficiency of the key components of our MeFlow. For all ablation studies, we train the models on FlyingChairs and FlyingThings3D, and evaluate on the Sintel and KITTI training set. For all experiments, memory and inference time are measured with 12 GRU-based iterations on our RTX 3090 GPU with 24G.

\begin{table}
\small
\caption{Comparisons with accuracy-oriented methods. For $1080\!\times\!1920$ resolution inputs, our method consumes $\mathbf{8 \times}$ \textbf{less memory} than GMA and SKFlow.}
    \centering 
    \setlength{\tabcolsep}{1.pt} %
    \begin{tabular}{lccccc}
    \toprule
    \multirow{2}{*}{Method } &\multirow{2}{*}{\begin{tabular}[x]{@{}c@{}}KITTI\\test \end{tabular}}
    &\multicolumn{2}{c}{$448 \times 1024$} & \multicolumn{2}{c}{$1080 \times 1920$} \\
    \cmidrule(lr){3-4} \cmidrule(lr){5-6}
    & & Mem.(G) & Time(ms) & Mem.(G) & Time(ms) \\
    \midrule
    % RAFT & 5.10 & 0.48 & 64 & 8.33 & 300 \\ 
    RAFT~\cite{raft} & 5.10 & \underline{0.48} & 64 & 8.33 & 300 \\
    GMA~\cite{gma} & 5.15 & 0.65 & 75 & 11.73 & 387 \\
    SepaFlow~\cite{separableflow} & \textbf{4.64} &0.65 &570 &12.13  &3948 \\
    GMFlow~\cite{gmflow} & 9.32 &1.31 & 115 & \underline{8.30} & 1242 \\
    SKFlow~\cite{skflow} & \underline{4.84} & 0.66 & 138 & 11.73 & 634 \\
    FlowFormer~\cite{flowformer} & 4.87 & 2.74 & 250 & OOM & - \\
    MeFlow (Ours) & 4.95 & \textbf{0.33} & 65 & \textbf{1.39} & 260 \\    
    \bottomrule
    \end{tabular}
    % , and FlowFormer~\cite{flowformer} causes an out-of-memory error (denoted as OOM).}
    \label{tab:comp_acc_ori}
\end{table}

Results of ablations are shown in \cref{tab:ablation}. Following RAFT, the Base denotes the 2D lookup on target image feature in each iteration, and the lookup radius $r$ is set to 4. The proposed OC (orthogonal correlation) can greatly reduce memory consumption and speed up inference, especially on high-resolution images. The OC only indexes $(2r\!+\!1)\!\times\!2$ target feature in two 1D orthogonal directions for each pixel on the source feature. To complement information loss, we exploit OA (orthogonal attention) to propagate the feature information of the 2D space to the orthogonal space, which can significantly improve the overall performance at a negligible cost. To efficiently provide information of large displacements, we propose a new radius-distribution multi-scale lookup strategy, named RDMS. The propose OC+OA+RDMS, denoted as MeFlow, achieves the best performance.

\begin{table}
\small
\caption{Analysis on horizontal ($h$) and vertical ($v$) cost volumes. EPE ($h$) and EPE ($v$) represent the end-point-error of the horizontal and vertical flow component, respectively.}
    \centering
    \begin{tabular}{llccc}
    \toprule
    \multirow{2}{*}{Cost Volume} & \multirow{2}{*}{Method} & \multicolumn{3}{c}{Sintel (train, clean)} \\
    \cmidrule(lr){3-5}
    & & EPE & EPE ($h$) & EPE ($v$) \\
    \midrule
    \multirow{2}{*}{$v$ attn, $h$ corr} & Flow1D & 3.10 & 1.66 & 2.12 \\
    & MeFlow & {\bf 1.79} & {\bf 1.14} & {\bf 1.10} \\
    \midrule
    \multirow{2}{*}{$h$ attn, $v$ corr} & Flow1D & 4.05 & 3.55 & 1.13 \\
    & MeFlow & {\bf 1.79} & {\bf 1.37} & {\bf 0.85} \\
    \midrule
    \multirow{2}{*}{concat both} & Flow1D & 1.98 & 1.48 & 0.94 \\
    & MeFlow & {\bf 1.49} & { \bf 1.08} & {\bf 0.79} \\    
    \bottomrule
    \end{tabular}
    \label{tab:cost_volume_hv}
    % \vspace{-10pt}
\end{table}

\begin{table}
\small
\caption{Comparisons with Flow1D in different ranges of optical flow regions. Our method performs better in large motion regions (s\textsubscript{10-40}).}
    \centering
    \begin{tabular}{lccc}
    \toprule
    \multirow{2}{*}{Method}  & \multicolumn{3}{c}{Sintel (test, final)} \\
    \cmidrule(lr){2-4}
    & EPE & s\textsubscript{0-10} & s\textsubscript{10-40}\\
    \midrule
    Flow1D &3.81 & 0.74 & 2.48\\ 
    Ours & \textbf{3.09} & \textbf{0.64} (14\%$\uparrow$) & \textbf{1.77} (29\%$\uparrow$)\\    
    \bottomrule
    \end{tabular}
    \label{tab:large_motion}
    \vspace{-10pt}
\end{table}

\begin{table}[t]
\small
\caption{Benchmark performance on Sintel and KITTI datasets. The numbers in parentheses are the results from fine-tuning the methods on the data.}
    \centering 
    \begin{tabular}{lcccccc}
    \toprule
    \multirow{2}{*}{Method }
    % &\multicolumn{2}{c}{Sintel (train)} 
    & \multicolumn{2}{c}{Sintel (test)} &\multicolumn{2}{c}{KITTI (F1-all)} \\
    \cmidrule(lr){2-3} \cmidrule(lr){4-5}
    & Clean & Final & train & test \\
    \midrule
    FlowNet2~\cite{flownet2} & 4.16 & 5.74 &(6.8) & 11.48 \\ 
    LiteFlowNet2~\cite{liteflownet2} & 3.48 & 4.69 &(4.8) & 7.74 \\ 
    PWC-Net+~\cite{pwc-net+} & 3.45 & 4.60 &(5.3) & 7.72 \\ 
    HD3~\cite{hd3}  & 4.79 & 4.67 &(4.1) & 6.55 \\ 
    IRR-PWC~\cite{irr-pwc} & 3.84 & 4.58 &(5.3) & 7.65 \\ 
    IRR-PWC-it~\cite{disentangling} & 2.19 & 3.55 & - & 5.73 \\ 
    VCN~\cite{vcn} & 2.81 & 4.40 &(4.1) & 6.30 \\ 
    DICL~\cite{dicl} & 2.12 & 3.44 &(3.6) & 6.31 \\
    MaskFlowNet~\cite{maskflownet} & 2.52 & 4.17 & - & 6.10 \\ 
    RAFT~\cite{raft} & 1.61 & 2.86 & (1.5) & 5.10 \\
    GMA~\cite{gma} &1.39 & 2.47 & (1.2) & 5.15 \\
    SepaFlow~\cite{separableflow} &1.50 & 2.67 & (1.6) & 4.64 \\  
    GMFlow~\cite{gmflow} & 1.74 & 2.90 & - & 9.32 \\
    GMFlow+~\cite{unimatch} & 1.03 & 2.37 & - & 4.49 \\
    SKFlow~\cite{skflow} &1.28 & 2.23 & (0.94) & 4.84 \\
    FlowFormer~\cite{flowformer} &1.18 & 2.36 & (1.1) & 4.87 \\
    DEQ-RAFT~\cite{deq} & 1.82 & 3.23 & (1.4) & 4.98 \\ 
    EMD-L~\cite{deng2023explicit} &1.32 &2.51 &- &4.51 \\
    SCV~\cite{scv} & 1.72 & 3.60 &(2.1) & 6.17 \\ 
    Flow1D~\cite{flow1d} & 2.24 & 3.81 &(1.6) & 6.27 \\ 
    MeFlow (Ours) & 2.05 & 3.09 &(1.6) & 4.95 \\ 
    \bottomrule
    \end{tabular}
    \label{tab:benckmark}
    \vspace{-10pt}
\end{table}

% \vspace{8pt}
% \noindent\textbf{Lookup radius.} We verify the effectiveness of the multi-scale lookup radius and show it in Tab. \ref{tab:radius_r}. The lookup radius $r\!=\!8$ at single resolution corresponds to 64 pixels in the original image resolution, while our multi-scale lookup radius $R\!=\!8$ corresponds to 128 pixels, which can model the correspondence of larger displacements when using the same computational cost. The proposed multi-scale (MS) lookup strategy achieves the best performance.

\begin{figure*}
\centering
\includegraphics[width=0.85\textwidth]{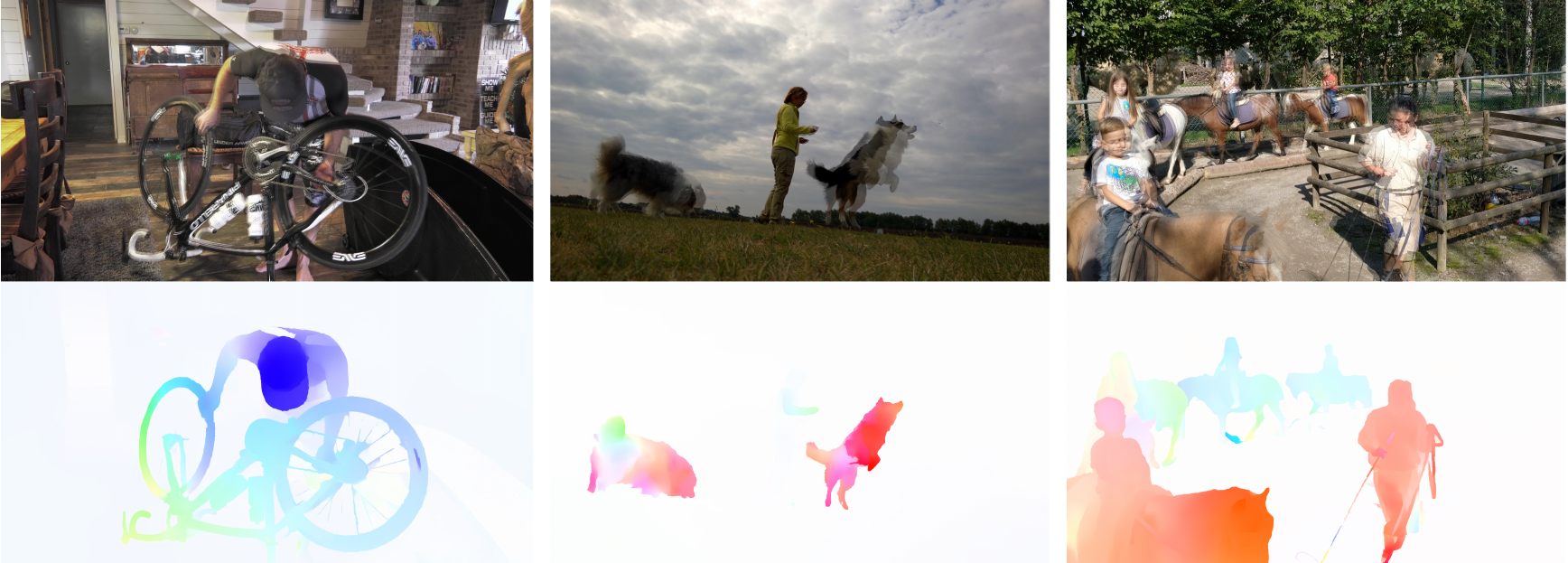}
\caption{Optical flow prediction results on 4K ($2160\times3840$) resolution images from DAVIS~\cite{davis17, davis19} datasets. From top to bottom: overlapping images and optical flow results.}
\label{fig:4k}
\vspace{-10pt}
\end{figure*}

\subsection{Comparison with Existing Methods}

\noindent\textbf{Comparison with existing cost volumes.} We conduct comprehensive comparisons with existing representative cost volume construction methods to demonstrate the superiority of our local orthogonal cost volume method. Results are shown in \cref{tab:compare_cost_volume}. FlowNet2~\cite{flownet2} constructs a single-scale cost volume, PWC-Net~\cite{pwc-net} constructs a coarse-to-fine cost volume pyramid, RAFT~\cite{raft} constructs an all-pairs 4D cost volume, and Flow1D~\cite{flow1d} constructs two 3D cost volumes. We evaluate the cross-dataset generalization ability on Sintel and KITTI training sets and show benchmark results on KITTI. In this evaluation, our method achieves the best performance in terms of accuracy and memory consumption. In particular, our method outperforms PWC-Net by 49.5\% and Flow1D by 21.1\% on KITTI, and exhibits higher memory efficiency. For high-resolution $1080 \times 1920$ images, our MeFlow can consume $6 \times$ less memory than RAFT.

\vspace{8pt}
\noindent\textbf{Comparison with memory-efficient methods.} We compare our MeFlow with memory-efficient methods based on recurrent optimization, shown in \cref{tab:comp_mem_effi}. Our method outperforms SCV~\cite{scv} and Flow1D~\cite{flow1d} on KITTI. Our method is slightly inferior compared to DIP~\cite{dip} in terms of accuracy but enjoys less memory consumption and inference time. Sequential ConvGRU in DIP is time-consuming, and our method yields a $2.4 \times$ speedup than DIP for $1080 \times 1920$ resolution images.

\vspace{8pt}
\noindent\textbf{Comparison with accuracy-oriented methods.} We compare our MeFlow with accuracy-oriented methods, shown in \cref{tab:comp_acc_ori}. Our method achieves competitive accuracy on the KITTI benchmark, and shows higher efficiency than these accuracy-oriented methods in terms of both memory consumption and inference speed. The superiority will be more significant for high-resolution images. For example, when the input image resolution is $1080 \times 1920$, our method consumes $\mathbf{8 \times}$ \textbf{less memory} than GMA~\cite{gma} and SKFlow~\cite{skflow}, and yields a $4.8\times$ speedup than GMFlow~\cite{gmflow}, $2.4\times$ speedup than SKFlow~\cite{skflow}. FlowFormer causes an out-of-memory error for $1080 \times 1920$ resolution on our 24G GPU.

\vspace{8pt}
\noindent\textbf{Analysis on horizontal and vertical cost volume.} We decompose the orthogonal cost volume into horizontal cost volume ($v$ attn, $h$ corr) and vertical cost volume ($h$ attn, $v$ corr) and evaluate the corresponding performance, shown in \cref{tab:cost_volume_hv}. With only horizontal or vertical cost volume, our method still achieves comparable performance, which is even better than Flow1D~\cite{flow1d} with both horizontal and vertical cost volumes. With the same cost volume setup, our method outperforms Flow1D by a large margin.

\vspace{5pt}
\noindent\textbf{High-resolution performance.} Our MeFlow can scale up well to very high-resolution images. As shown in \cref{fig:comp_2k}, we show visual comparisons on the high-resolution ($1080\!\times\!1920$) DAVIS dataset. Our method achieves comparable results with RAFT while consuming $6\times$ less memory, and significantly better results than Flow1D while consuming similar memory. In \cref{fig:4k}, we also show visual results on 4K ($2160\!\times\!3840$) resolution images from DAVIS datasets. Our MeFlow can handle 4K images with only 5.4G memory usage. RAFT can not handle 4K images due to causing an out-of-memory error on our 24G GPU. 

% \paragraph{Memory efficiency analysis.} 
% We measure the practical memory consumption under different input resolutions in Figure \ref{fig:mem_effi_ana}. As the input resolution increases, the memory consumption of our method grows nearly linearly. While RAFT will quickly cause an out-of-memory error on our 24G GPU. 
% For the input feature map of size $H\!\times\!W\!\times\!D$, the computational complexity of constructing an all-pairs 4D cost volume in RAFT~\cite{raft} is $\mathcal{O} ((HW)^2 D)$, and the computational complexity of constructing two 3D cost volumes in Flow1D~\cite{flow1d} is $\mathcal{O} (HW(H+W)D)$. As a comparison, our multi-scale orthogonal cost volume is $\mathcal{O} (HW(4R+1)D)$ ($R=8$), where $(4R\!+\!1) \ll H\!+\!W \ll HW$ for high-resolution inputs.

\subsection{Benchmark Results}
The results of our MeFlow on Sintel and KITTI benchmarks are shown in \cref{tab:benckmark}. Our method achieves comparable results with other methods. On Sintel and KITTI benchmark, our method is inferior to some accuracy-oriented methods such as SKFlow~\cite{skflow} and FlowFormer~\cite{flowformer}, but outperforms most existing methods such as PWC-Net+~\cite{pwc-net+}, MaskFlowNet~\cite{maskflownet} and Flow1D~\cite{flow1d}. Specially, our method outperforms the equally memory-efficient Flow1D by a large margin, e.g., exceeding Flow1D 21.1\% on KITTI and 18.9\% on Sintel (Final). We also show visual comparisons on the Sintel and KITTI test sets, shown in \cref{fig:sintel} and \cref{fig:kitti}. Compared with Flow1D, our method can perform better in the fine structure and large textureless areas. Specially, Flow1D struggles to handle large motions since its two 3D cost volumes lose matching information for large displacements, as illustrated in \cref{fig:large_motion} and \cref{tab:large_motion}. In contrast, our method can handle large motions (s\textsubscript{0-10}) well.

%% file: sec/5_conclusion.tex
\section{Conclusion}
\label{sec:conclusion}

The full 4D cost volume in RAFT and global matching in GMFlow achieve striking performance for optical flow estimation. However, their memory consumption grows quadratically with input resolution, making them impractical for high-resolution images. In this paper, we propose a novel local orthogonal cost volume that decomposes the 2D search space dynamically into two 1D orthogonal spaces by attention, enabling our method to scale well to high-resolution images. We further propose a new radius-distribution multi-scale lookup strategy to model the correspondences of large displacements. Our MeFlow achieves the highest memory efficiency among all methods and competitive accuracy compared to the state-of-the-art. We hope the proposed efficient local orthogonal cost volume will inspire future research on high-resolution optical flow estimation.